\documentclass[times, review, 10pt]{elsarticle}

\usepackage{amssymb}
\usepackage{amsmath}
\usepackage{hyperref}
\usepackage{xcolor}
\usepackage[table]{xcolor}
\usepackage{xspace}
\usepackage[inkscapeformat=png]{svg}
\usepackage{subcaption}
\usepackage{caption}
\usepackage{graphicx}
\usepackage{multirow}
\usepackage{booktabs}
\usepackage{rotating}
\usepackage{makecell}

\newcommand{\eg}{\textit{e.g}.\xspace}
\newcommand{\ie}{\textit{i.e.}\xspace}
\newcommand{\etc}{\textit{etc.}\xspace}
\newcommand{\DatabaseName}[0]{AVAPrintDB}

\journal{Pattern Recognition}

\begin{document}

\begin{frontmatter}



\title{Leveraging Avatar Fingerprinting: A Multi-Generator Photorealistic Talking-Head Public Database and Benchmark} 


\author[UAM]{Laura Pedrouzo-Rodriguez} 
\author[UAM]{Luis F. Gomez} 
\author[UAM]{Ruben Tolosana}
\author[UAM]{\\Ruben Vera-Rodriguez}
\author[UAM]{Roberto Daza}
\author[UAM,ULPGC]{Aythami Morales}
\author[UAM]{Julian Fierrez}

\affiliation[UAM]{organization={BiometricsAI},
            addressline={Universidad Autónoma de Madrid}, 
            city={Madrid},
            postcode={28049}, 
            country={Spain}}
\affiliation[ULPGC]{organization={Department of Mathematics},
            addressline={Universidad de Las Palmas de Gran Canaria}, 
            city={Las Palmas de Gran Canaria},
            postcode={35001}, 
            country={Spain}}

\begin{abstract}
Recent advances in photorealistic avatar generation have enabled highly realistic talking-head avatars, raising security concerns regarding identity impersonation in AI-mediated communication. To advance this challenging problem, the task of avatar fingerprinting aims to determine whether two avatar videos are driven by the same human operator. However, current public databases in the literature are scarce and based solely on old-fashioned talking-head avatar generators, not representing realistic scenarios for the current task of avatar fingerprinting. To overcome this situation, the present article introduces \DatabaseName, a new publicly available multi-generator talking-head avatar database for avatar fingerprinting. \DatabaseName~is constructed from two audiovisual corpora and three state-of-the-art avatar generators (GAGAvatar, LivePortrait, HunyuanPortrait), representing different synthesis paradigms, and includes both self- and cross-reenactments to simulate legitimate usage and impersonation scenarios.

Building on this database, we define a standardized and reproducible benchmark for avatar fingerprinting, considering public state-of-the-art avatar fingerprinting systems and exploring novel methods based on Foundation Models (DINOv2 and CLIP). We also conduct a comprehensive analysis under generator and dataset shift. Our results show that, while identity-related motion cues persist across synthetic avatars, current avatar fingerprinting systems remain highly sensitive to changes in the synthesis pipeline and source domain. The \DatabaseName, benchmark protocols, and avatar fingerprinting systems are publicly available to facilitate reproducible research.
\end{abstract}




\end{frontmatter}



\section{Introduction}\label{sec:intro}

Recent progress in photorealistic avatar generation has enabled the synthesis of highly realistic talking-head avatars, significantly advancing applications in telepresence~\cite{jang2025exploring}, virtual reality~\cite{FRASER2024100082}, digital humans~\cite{qu2025humanoid}, and AI-mediated communication~\cite{mandic2025teachers}. Large technology companies and startups are actively investing in avatar-based interfaces\footnote{\href{https://www.theguardian.com/technology/2026/jan/26/uk-ai-startup-synthesia-almost-doubles-valuation-4bn-funding-round-corporate-video-avatars}{UK maker of AI avatars nearly doubles valuation to \$4bn after funding round}, \href{https://www.vogue.com/article/fashion-metaverse-insights-startup-geeiq-raises-dollar85-million?}{Fashion metaverse insights startup Geeiq raises \$8.5 million}}, with applications ranging from virtual assistants and customer service agents to remote collaboration platforms and personalized digital identities~\cite{mandic2025teachers}. Beyond content creation, photorealistic avatars are increasingly envisioned as persistent digital representatives of users in real-time interactions. In such settings, avatars can preserve privacy, enable language adaptation, or provide a consistent identity across platforms. For example, a remote employee may attend a videoconference using a personalized avatar, a clinician may communicate through a digital human interface, or a customer-facing assistant may operate via a branded synthetic persona. As avatars become active participants in communication and decision-making processes, securely establishing \emph{who controls the avatar} becomes a critical requirement for authentication, accountability, and trust in AI-mediated systems. Industry reports\footnote{\href{https://www.marketsandmarkets.com/Market-Reports/ai-avatar-market-146528536.html}{Digital Human Avatar Market Size Report (Markets and Markets, 2025)}, \href{https://www.mckinsey.com/capabilities/growth-marketing-and-sales/our-insights/value-creation-in-the-metaverse}{Value Creation in the Metaverse (McKinsey \& Company, 2023)}} estimate that the global market for digital humans and avatar-based communication systems is expected to grow rapidly in the coming years, driven by advances in generative AI and increasing adoption of immersive communication technologies.

As avatar-mediated communication becomes increasingly integrated into real-world digital ecosystems, the ability to reliably authenticate avatar operators becomes critical not only for usability and personalization, but also for preventing impersonation, fraud, and misuse, mitigating new security and privacy risks. Recent incidents have demonstrated that synthetic media can be exploited for impersonation attacks, social engineering, and identity fraud. For example, DeepFake-based scams\footnote{\href{https://www.europol.europa.eu/cms/sites/default/files/documents/Europol_Innovation_Lab_Facing_Reality_Law_Enforcement_And_The_Challenge_Of_Deepfakes.pdf}{Facing reality? Law enforcement and the challenge of deepfakes (Europol Innovation Lab, 2024)}} have been used to impersonate executives in video calls and fraudulently authorize financial transactions, leading to losses of millions of dollars in real-world cases\footnote{\href{https://edition.cnn.com/2024/02/04/asia/deepfake-cfo-scam-hong-kong-intl-hnk}{Finance worker pays out \$25 million after video call with deepfake `chief financial officer'}, \href{https://www.bbc.com/news/articles/c0j59vydxj9o}{Deepfake attack: 'Many people could have been cheated'}}. As avatar-based communication becomes more prevalent, these risks extend beyond static DeepFakes to interactive real-time impersonation scenarios, where an attacker can control a synthetic avatar that visually appears to be a legitimate user.

Traditional authentication methods based on appearance, such as face recognition and DeepFake detection, address problems that are related to, but fundamentally different from, the one considered in this work. In biometric face recognition~\cite{DEANDRESTAME2025103099}, the objective is to verify the identity from appearance cues extracted from facial images or videos. The visual identity itself constitutes the biometric evidence. In contrast, the appearance cannot serve as reliable evidence in avatar authentication, since the avatar face may correspond to an arbitrary target identity selected by the user, an impersonator, or the platform. DeepFake detection~\cite{TOLOSANA2020131,handbookDFc,9115874}, on the other hand, addresses a different question: determining whether a media sample is real or synthetically generated. Its objective is authenticity verification, often by detecting synthesis artifacts or inconsistencies. The avatar authentication task considered in this work is more challenging, as all samples are already synthetic avatars.

\begin{figure}[t]
    \centering
    \includegraphics[width=\linewidth]{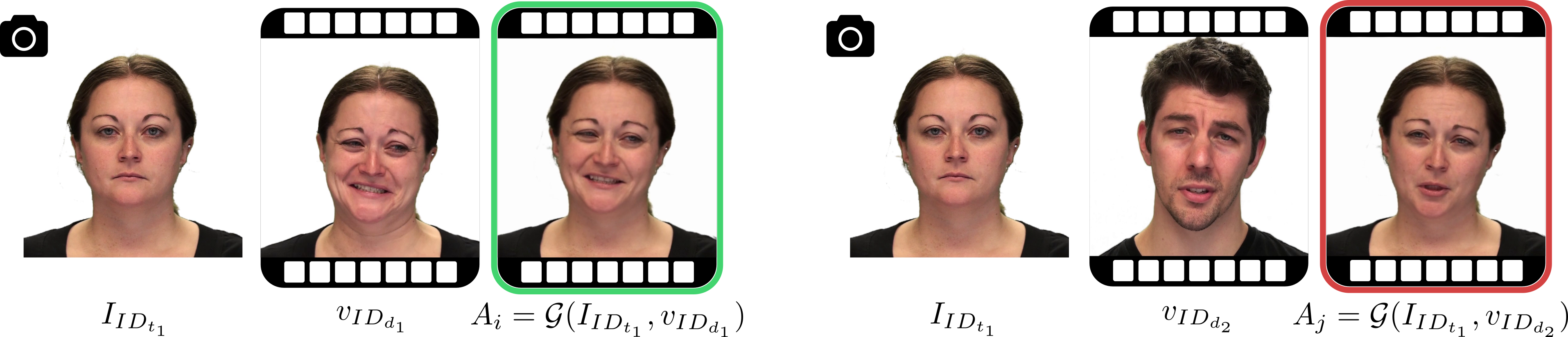}
    \caption{\textbf{Genuine and impostor avatars}. 
    \textit{Left}: \textcolor{green}{Genuine Avatar}: target identity $ID_{t_1}$ is the same as the driving identity, $ID_{d_1}$. \textit{Right}: \textcolor{red}{Impostor Avatar}: target identity $ID_{t_1}$ and driving identity $ID_{d_2}$ are different.}
    \label{fig:avatar-fingerprinting-concept}
\end{figure}

This shift fundamentally changes the authentication problem: instead of verifying \emph{who the avatar looks like}, the system must determine \emph{who is controlling the avatar}. This authentication task is known in the literature as \emph{avatar fingerprinting}~\cite{prashnani2024avatar}. Formally, avatar fingerprinting can be formulated as a pairwise biometric verification problem operating in the synthetic domain. Let an avatar video be represented as $A = \mathcal{G}(I_{ID_t}, v_{ID_d})$ where $\mathcal{G}$ denotes an avatar generator, $ID_t$ is the \textit{target} identity defining the avatar appearance, represented as an image $I$, and $ID_d$ is the \textit{driving} identity  providing the driving signal $v_{ID_d}$ (\eg, facial motion and head dynamics extracted from a real video), as depicted in Fig.~\ref{fig:avatar-fingerprinting-concept}. The objective is to determine whether two avatar videos $A_1$ and $A_2$ (\ie, a genuine video available in the gallery and a doubtful test video) are controlled by the same underlying driving identity, $ID_d$, \ie estimating $H_0 : ID_{d_1} = ID_{d_2}$ versus $H_1 : ID_{d_1} \neq ID_{d_2}$, where $ID_{d_1}$ and $ID_{d_2}$ denote the driving identities associated with avatar videos $A_1$ and $A_2$. 

Despite the considerable security implications for society~\cite{daza2026ares,pavel25iccv}, avatar fingerprinting is a relatively new and barely explored topic. Existing studies presented in the literature~\cite{prashnani2024avatar,alsherfawi2025decoding,pedrouzo2025really} suffer from limited reproducibility and insufficient evaluation under realistic up-to-date conditions. Neither~\cite{prashnani2024avatar} nor~\cite{alsherfawi2025decoding} have publicly released their avatar fingerprinting system. Furthermore, current datasets are restricted in scope \eg, the popular NVFAIR dataset~\cite{prashnani2024avatar} relies on old-fashioned reenactment techniques (FaceVid2Vid~\cite{wang2021one}, TPS~\cite{zhao2022thin}, LIA~\cite{wang2022latent}) and is only available upon request. As a result, it is difficult to advance in this challenging task and know whether current avatar fingerprinting systems truly capture identity-dependent motion cues or instead rely on artifacts specific to a given dataset or synthesis pipeline.

This limitation is particularly critical in light of the rapid evolution of avatar generation technologies. Modern generators span multiple paradigms (\eg, warping-based models~\cite{Wang_2025_CVPR}, 3D Gaussian splatting~\cite{qian20233dgsavatar,yan20252dgsavatar}, diffusion-based approaches~\cite{zhang2025layeravatar}, etc.), each introducing different motion transfer characteristics, temporal smoothing behaviors, and artifact profiles. In real-world scenarios, an avatar fingerprinting system is likely to encounter avatar videos generated by unseen AI models and under various conditions. Therefore, evaluating robustness across generators and datasets is essential to assess the practical viability of avatar fingerprinting.

To address these limitations and advance reproducible research in avatar fingerprinting, the present work makes the following contributions:

\begin{itemize}
    \item \textbf{First Fully Public Multi-Generator Avatar Fingerprinting Database}: We introduce \DatabaseName\footnote{\label{fn:github}\url{https://github.com/BiometricsAI/AVAPrintDB}}, the first fully public avatar fingerprinting resource combining a large-scale database, standardized benchmark protocols, and publicly available systems/code. The benchmark comprises more than 66,000 avatar videos generated from two audiovisual source datasets (CREMA-D~\cite{cao2014crema} and RAVDESS~\cite{livingstone2018ryerson}) and three state-of-the-art avatar generators (GAGAvatar~\cite{chu2024generalizable}, LivePortrait~\cite{guo2024liveportrait} and HunyuanPortrait~\cite{hunyuan2025cvpr}), representing distinct synthesis paradigms and including both self- and cross-reenactments simulating legitimate usage and impersonation attacks).
    \item \textbf{Systematic Multi-Generator and Cross-Domain Benchmarking}: We define a standardized and reproducible evaluation framework for avatar fingerprinting, including identity-disjoint splits, predefined verification trials, and evaluation protocols. The benchmark explicitly evaluates robustness under generator shift, dataset shift, and demographic variability, enabling realistic assessment of domain generalization in avatar fingerprinting.
    \item \textbf{First Exploration of Foundation Models for Avatar Fingerprinting}: To the best of our knowledge, we investigate for the first time the use of DINOv2~\cite{oquab2024dinov} and CLIP~\cite{radford2021learning} as feature extractors for avatar fingerprinting. Combined with a temporal aggregation mechanism, we analyze their ability to capture identity-dependent behavioral cues and compare them against landmark-based approaches.
\end{itemize}

The article is organized as follows. Sec.~\ref{sec:relatedwork} provides a revision of the state of the art. Sec.~\ref{sec:db} describes the details of our proposed \DatabaseName. Sec.~\ref{sec:benchmark} describes the proposed benchmark, including the experimental protocol and the public avatar fingerprinting systems. The experimental results are included in Sec.~\ref{sec:experiments}. Finally, Sec.~\ref{sec:conclusions} concludes with key takeaways.

\section{Related Work}\label{sec:relatedwork}

\subsection{Photorealistic Talking-Head Avatar Generation}\label{ssec:avatargenerators}

Photorealistic talking-head avatar generation aims to animate a target identity using the facial motion and head pose extracted from a driving signal, typically a video sequence. This task is also commonly referred to as \emph{portrait animation}~\cite{cui2025hallo3}, \emph{motion transfer}~\cite{SUN2024104138}, or \emph{talking-head reenactment}~\cite{11209919}. It is important to note that this task differs from related problems such as \emph{novel-view synthesis}~\cite{sun2023next3d} or \emph{audio-driven animation}~\cite{info15110675}, as the goal is to preserve the appearance of a reference identity while faithfully transferring motion dynamics from another source.

Early approaches to facial reenactment relied on parametric face models and 3D morphable models to retarget expressions and head pose, enabling real-time portrait manipulation and reenactment~\cite{thies2016face2face}. The emergence of deep generative models allowed realistic reenactment from minimal input, particularly through motion transfer methods that animate a single portrait using learned keypoint dynamics~\cite{wang2021one,siarohin2019first}. Subsequent advances in neural rendering and implicit scene representations improved realism and view consistency~\cite{mildenhall2021nerf,grassal2022neural}, while recent diffusion-based video generation techniques have further increased photorealism and temporal coherence~\cite{hunyuan2025cvpr,ho2022video}. 

These methods can be broadly categorized according to their underlying representation and synthesis strategy. Neural rendering approaches achieve high visual fidelity and view consistency, but are typically computationally expensive and often require multiple view data~\cite{mildenhall2021nerf, kirschstein2023nersemble}. Gaussian splatting methods offer a more efficient alternative with real-time capabilities, although they may exhibit some geometric inconsistencies~\cite{qian2024gaussianavatars,xiang2024flashavatar,ren2026omg}. Image-based warping methods are lightweight and practical, and recent advances have demonstrated strong visual quality despite the lack of explicit 3D modeling~\cite{guo2024liveportrait}. Finally, diffusion-based methods represent the current state of the art in realism and temporal coherence, at the cost of higher computational requirements~\cite{hunyuan2025cvpr}.

Across these paradigms, different design choices lead to variations in motion transfer fidelity, temporal consistency, and artifact characteristics. These differences are particularly relevant for avatar fingerprinting, as they can influence how identity-related facial dynamics are preserved or distorted. Consequently, evaluating avatar fingerprinting systems across multiple avatar generation methods is essential to assess their robustness under realistic synthesis variability, which is the purpose of this article.

\begin{table}[t]
\centering
\renewcommand{\arraystretch}{0.7}
\setlength{\tabcolsep}{4pt}
\begin{tabular}{lccc}
\toprule
\textbf{Property} &
\textbf{NVFAIR\cite{prashnani2024avatar}} &
\textbf{GAGAvatarDB\cite{pedrouzo2025really}} &
\textbf{\DatabaseName} \\
\midrule

Public access &
Under request &
Under request &
Yes \\

Public benchmark protocol &
Limited &
Yes &
Yes \\

Public systems / code &
No &
Yes &
Yes \\

Multiple generators &
Yes (legacy) &
No &
Yes (modern) \\

Cross-domain evaluation &
Limited &
Limited &
Yes \\

Generator diversity &
Moderate &
Low &
High \\
\bottomrule
\end{tabular}
\caption{\textbf{Comparison between existing avatar fingerprinting resources and the proposed \DatabaseName.}}
\label{tab:benchmark_comparison}
\end{table}

\subsection{Avatar Fingerprinting Databases and Benchmarks}\label{ssec:databases}

To date, public resources for avatar fingerprinting remain scarce and fragmented. Table~\ref{tab:benchmark_comparison} provides a comparison of our proposed \DatabaseName~with existing resources in the literature. As can be seen, they differ in accessibility, reproducibility, generator diversity, and evaluation support. 

The NVFAIR database introduced by NVIDIA~\cite{prashnani2024avatar} is currently the largest collection of reenacted talking-head videos, containing more than 600,000 videos. However, access is restricted through the request form, and the benchmark relies on older reenactment methods (FaceVid2Vid~\cite{wang2021one}, TPS~\cite{zhao2022thin}, and LIA~\cite{wang2022latent}), which no longer represent the current state of photorealistic avatar generation, as discussed in Sect.~\ref{ssec:avatargenerators}. Also, benchmarking is not possible because the proposed implementation of the avatar fingerprinting system is not publicly available.

To our knowledge, the only fully public benchmark prior to this work is the GAGAvatar-Benchmark~\cite{pedrouzo2025really}. However, it only covers a single talking-head avatar generator (GAGAvatar~\cite{chu2024generalizable}), limiting diversity in appearance and motion dynamics. 

In contrast, the proposed \DatabaseName~and benchmark provides a fully public benchmark including the database, generation details, evaluation protocols, and avatar fingerprinting systems. It combines three up-to-date photorealistic avatar generators spanning different synthesis paradigms (GAGAvatar~\cite{chu2024generalizable}, LivePortrait~\cite{guo2024liveportrait} and HunyuanPortrait~\cite{hunyuan2025cvpr}), two source datasets, identity-disjoint protocols, and systematic evaluation under generator and dataset shift. These characteristics enable reproducible benchmarking under more realistic and diverse conditions than previous available resources.

\subsection{Avatar Fingerprinting Systems}\label{ssec:avatarverification}

The task of avatar fingerprinting is formulated as a pairwise decision problem: given two avatar videos, the goal is to determine whether \textit{they are driven} by the same underlying identity or not. This differs from authentication, which evaluates a sample against a claimed identity (\eg, access control scenarios), and from identification, where a sample is matched against a gallery of known identities. Avatar fingerprinting constitutes the most fundamental setting, as it isolates the ability of a model to capture identity-dependent behavioral signals without relying on predefined identity classes.

Early work on avatar fingerprinting demonstrated that such behavioral cues persist through avatar generation. In~\cite{prashnani2024avatar}, identity was modeled using facial landmark dynamics using temporal convolutional networks, showing that user-specific motion patterns can be discriminative even in highly realistic avatars (they achieved around 87\% AUC in their best setting). Similarly,~\cite{alsherfawi2025decoding,pedrouzo2025really} proposed a graph-based approach that captures structured relationships between facial landmarks over time, obtaining similar results in their respective experiments (86.6\% and 83\% AUC, respectively). 

A common characteristic of existing approaches is their reliance on landmark-based representations, which provide invariance to appearance and mitigate the influence of rendering artifacts. However, this design choice may also limit the ability to capture fine-grained visual cues present in the pixel domain, particularly those related to texture-dependent motion or subtle temporal inconsistencies. For this reason, in the present work, we explore for the first time the use of Foundation Models (DINOv2~\cite{oquab2024dinov} and CLIP~\cite{radford2021learning}) for avatar fingerprinting, which can encode richer spatio-temporal information without explicit geometric modeling.

\section{\DatabaseName: Proposed Avatar Fingerprinting Database}\label{sec:db}

\subsection{Real Databases: Source Videos}\label{sssec:source_database}

We selected two widely used video datasets in the literature that consider realistic scenarios for avatar fingerprinting: CREMA-D~\cite{cao2014crema} and RAVDESS~\cite{livingstone2018ryerson}. 
Our selection is guided by three main considerations. First, both datasets are distributed under permissive licenses that allow redistribution and derivative use, making them suitable for a reproducible benchmark. Second, they have been used in previous avatar fingerprinting and behavioral biometric work~\cite{prashnani2024avatar,pedrouzo2025really,vahdati2025unmasking}, facilitating continuity. Third, they include variations in emotional expression and scripted utterances, allowing controlled protocols in which different subjects produce identical sentences with matched affect. 
Such control lets identity-related motion cues be isolated from linguistic content, affective expression, and other confounding factors (\eg, comparing different individuals speaking the same sentence, or the same individual rendered through different pipelines).

We also considered the NVFAIR dataset~\cite{prashnani2024avatar}, which adds videoconferencing recordings from 46 participants, but its license restricts redistribution and limits demographic analysis, so we could not include it. 

\subsubsection{CREMA-D Dataset}
CREMA-D~\cite{cao2014crema} (Crowd-sourced Emotional Multimodal Actors Dataset) is a large-scale audiovisual corpus designed to study emotional speech and multimodal expression. It contains 7,438 video recordings of 91 actors (48 male, 43 female) between 20 and 74 years of age, representing a demographically diverse group with respect to race and ethnicity. The actors were recorded in controlled indoor conditions with frontal viewpoints, consistent lighting, and high-quality audio recording.

Each participant delivers 12 semantically neutral sentences, performed using multiple emotional expressions: anger, disgust, fear, happiness, neutral, and sadness. The recordings exhibit a natural variation in intensity and articulation, while preserving consistent linguistic content. All videos were recorded in $460\times360$ pixels at 30 FPS.

\subsubsection{RAVDESS Dataset}

RAVDESS~\cite{livingstone2018ryerson} (Ryerson Audio-Visual Database of Emotional Speech and Song) is a high-quality audiovisual corpus designed for research on emotional expression and multimodal speech. It includes recordings from 24 professional actors (12 female, 12 male) speaking and singing with a range of emotional expressions. In this article, we use the speech subset, which contains 1,440 video recordings corresponding to two repetitions of 60 scripted utterances produced by each actor.

The recordings were captured in tightly controlled studio conditions, including uniform lighting, frontal framing, neutral backgrounds, and high-fidelity audio and video. Actors deliver semantically neutral sentences using multiple emotional states (neutral, calm, happy, sad, angry, fearful, surprised, and disgust) with two intensity levels for all emotions but neutral. All videos are $1280\times720$ pixels and were recorded at 30FPS.

\subsubsection{Post-Processing Stage}\label{sssec:curated}

We discard the four videos reported by the creators of CREMA-D~\cite{cao2014crema} as corrupted, following the recommendations in their official repository. We further filter the dataset to retain only the 85 identities that contain at least 72 valid samples each. This results in a subset of 6,120 videos, ensuring a minimum number of samples per identity and reducing class imbalance. In addition, where metadata are available, the selection is stratified to preserve diversity in gender, age ranges, and ethnicity.

Regarding RAVDESS~\cite{livingstone2018ryerson}, we consider the entire dataset, as it provides a balanced distribution across identities and recording conditions. Only the speech subset is retained, excluding singing sequences to maintain consistency with CREMA-D.

RAVDESS provides the gender of the recorded actors. For ethnicity estimation, FaRL~\cite{zheng2022general} (a CLIP-based model) was used to classify into the same groups as in CREMA-D. For age estimation, GPT-4 was used.

An important difference between the two datasets lies in the temporal duration of the videos. CREMA-D videos are generally shorter, with most sequences containing between 2 and 3 seconds, whereas RAVDESS videos are 3 to 4 seconds long. Consequently, models trained or evaluated on these datasets may exhibit different performance characteristics depending on their ability to capture and aggregate temporal information.
The summary of the curated datasets is presented in Table~\ref{tab:curated_summary}. This curated video corpus is used as the source data for the avatar generation described in Sec.~\ref{ssec:pairs_pipeline}.

\begin{table}
\centering
\renewcommand{\arraystretch}{0.7}
\setlength{\tabcolsep}{3pt}
\begin{tabular}{ccccc}
\textbf{Dataset} & \textbf{\#IDs} & \textbf{\#Vids/ID} & \textbf{\#Videos} & \textbf{\#Vids/statement}\\ 
\hline
CREMA-D~\cite{cao2014crema} & 85 & 72 & 6,120 & 510 \\
RAVDESS~\cite{livingstone2018ryerson} & 24 & 60 & 1,440 & 720 \\
\hline
TOTAL & 109 & - & 7,560 & - \\
\hline
\end{tabular}
\caption{\textbf{Summary of the curated source datasets used to generate the avatar videos.}}
\label{tab:curated_summary}
\end{table}

\subsection{Photorealistic Talking-Head Avatar Generators}\label{ssec:avatar_generation}

A key aspect of our proposed \DatabaseName~is the inclusion of multiple state-of-the-art avatar generators, each relying on different modeling assumptions and motion transfer mechanisms: one from the Gaussian splatting family (GAGAvatar~\cite{chu2024generalizable}), one from the warping-based family (LivePortrait~\cite{guo2024liveportrait}), and one from the diffusion-based family (HunyuanPortrait~\cite{hunyuan2025cvpr}). These three generators represent distinct synthesis paradigms and therefore introduce different types of motion transfer artifacts that may influence the avatar fingerprinting task, and were selected for their public availability, efficiency, and visual quality. GAGAvatar~\cite{chu2024generalizable} is a one-shot 3D Gaussian head avatar method that reconstructs a controllable 3D head from a single image in a single forward pass, lightweight in GPU memory and offering real-time reenactment speeds. LivePortrait~\cite{guo2024liveportrait} provides a fast and efficient 2D animation solution with minimal memory overhead and qualitatively strong visual results. Finally, HunyuanPortrait~\cite{hunyuan2025cvpr} is a state-of-the-art diffusion model that, although heavier, produces high visual fidelity and temporal consistency, allowing us to evaluate a high-quality diffusion approach against faster lightweight methods.

From a biometric behavioral perspective, the generation process can be interpreted as a transformation that maps real facial motion to a synthetic domain. Ideally, this transformation should preserve identity-specific temporal patterns while altering the appearance. However, in practice, avatar generators may introduce biases, artifacts, or normalization effects that partially suppress or modify these cues. As a result, models trained or evaluated on a single generator may inadvertently learn generator-specific artifacts instead of robust behavioral patterns. To address this limitation, our proposed benchmark incorporates multiple generators with diverse characteristics, enabling systematic evaluation under intra- and cross-generator conditions. This design allows us to analyze whether avatar fingerprinting systems generalize across synthesis pipelines or rely on artifacts tied to a specific generator. 

In addition to synthesis characteristics, practical deployment considerations, such as computational cost, may also influence generator selection in real-world systems. Therefore, we report in our \href{https://github.com/BiometricsAI/AVAPrintDB}{GitHub} representative complexity indicators for the considered avatar generators, including approximate inference runtime, GPU memory footprint, and model scale under our experimental setup. These measurements are intended to provide operational context rather than constitute a dedicated efficiency benchmark.

\subsection{\DatabaseName~Generation and Analysis}\label{ssec:pairs_pipeline}

\begin{figure}[t]
    \centering
    \includegraphics[width=0.92\linewidth]{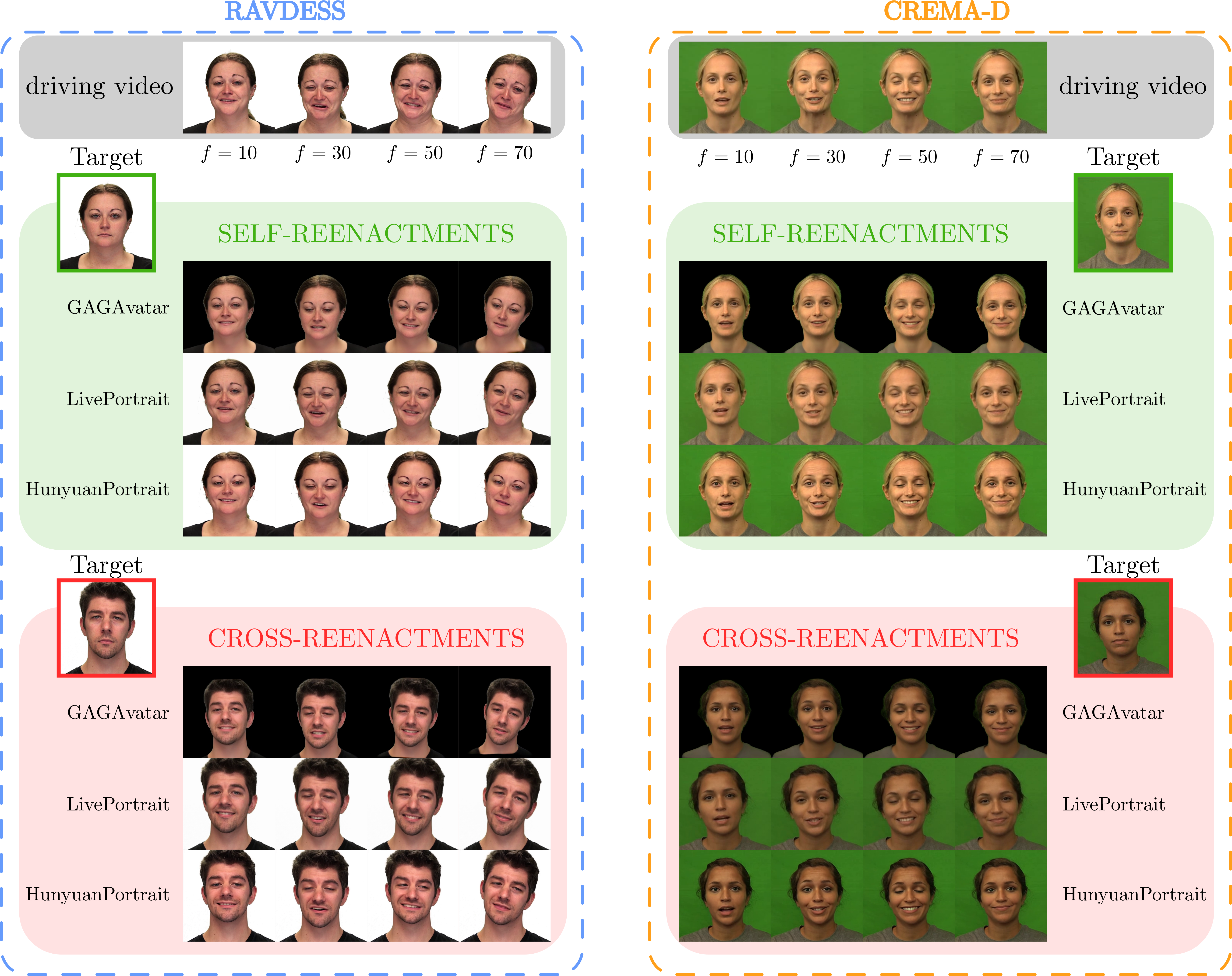}
    \caption{\textbf{Visualization of \textit{self-} and \textit{cross-reenactment} examples for each avatar generator and dataset.}}
    \label{fig:avatars_examples}
\end{figure}

Based on the curated datasets described in Sec.~\ref{sssec:source_database} and the avatar generators described in Sec.~\ref{ssec:avatar_generation}, we construct our proposed \DatabaseName~designed to support a controlled and reproducible evaluation of avatar fingerprinting systems. Each generated video is defined by a \emph{target identity} (which determines the appearance of the avatar) and a \emph{driving identity} (which provides the motion dynamics). This setup enables both \emph{self-reenactment} (same target and driver) and \emph{cross-reenactment} (different target and driver), as shown in Fig.~\ref{fig:avatars_examples}, which are essential for evaluating avatar fingerprinting. Every target-driver pair is rendered with the same three avatar generators, ensuring that each underlying motion instance is available under three different synthesis conditions. In addition, all avatar videos are generated using a consistent protocol. For the target images, we manually select a single frontal, neutral-expression frame per subject from the curated dataset. Manual selection ensures that the chosen frames are free of occlusions, head rotations, or strong expressions that could introduce artifacts during reenactment. Fig.~\ref{fig:avatars_examples} shows visual examples of avatars generated using the three selected methods, illustrating the process of generating self-reenactments and cross-reenactments across different time frames in the video. We next describe the specific details for building the self- and cross-reenactment pairs of our proposed \DatabaseName.

For each identity $\textrm{ID}_i$, we generate self-reenactments by pairing its target image with its own driving videos. In the driving video $v_{({\textrm{ID}_i,k)}}$ of identity $\textrm{ID}_i$, we generate an avatar sequence $A^{(g)}_{\textrm{ID}_i \leftarrow (\textrm{ID}_i,k)} = \mathcal{G}^{(g)}(I_{\textrm{ID}_i}, v_{(\textrm{ID}_i,k)})$, where $I_{\textrm{ID}_i}$ is the target image of identity $\textrm{ID}_i$,  $k$ indexes the driving video and $\mathcal{G}^{(g)}$ denotes generator $g \in \{$GAGA, LIVE, HUNY$\}$, where GAGA, LIVE and HUNY correspond to GAGAvatar~\cite{chu2024generalizable}, LivePortrait~\cite{guo2024liveportrait}, and HunyuanPortrait~\cite{hunyuan2025cvpr}, respectively. This yields, for every original driving video, three synthetic counterparts that share the same underlying facial motion but differ in synthesis artifacts and rendering style (generator-specific).

To simulate impersonation, for each identity $\textrm{ID}_i$, we select a fixed set of eight \emph{distinct target identities}, denoted $\mathcal{T}(\textrm{ID}_i) = \{\textrm{ID}_1, \textrm{ID}_2, \ldots, \textrm{ID}_j\}$ by $\textrm{ID}_j \neq \textrm{ID}_i$. Intuitively, the identity $\textrm{ID}_i$ acts as the \emph{driver} (operator) and the selected identities $\textrm{ID}_j$ act as the \emph{avatar appearance} (claimed identity). The targets are selected according to a deterministic sampling procedure designed to maximize target diversity (\eg, four males and four females) while preserving reproducibility. The selected targets remain constant on all driving videos with the same driver identity. The complete target-selection metadata and generation configuration files are publicly available in our \href{https://github.com/BiometricsAI/AVAPrintDB}{GitHub}.

Given the complete set of driving videos available for $\textrm{ID}_i$, we randomly sample (without replacement) $N$ driving videos $\{v_{(\textrm{ID}_i,k)}\}_{k \in [1,N ]}$ for cross-reenactments. The value of $N$ is dataset-dependent, determined according to the variability of each dataset, including factors such as emotional diversity and the range of statements available. For each target $\textrm{ID}_j \in \mathcal{T}(\textrm{ID}_i)$ and each sampled driving video $v_{(\textrm{ID}_i,k)}$, we generate $A^{(g)}_{\textrm{ID}_j \leftarrow (\textrm{ID}_i,k)} = \mathcal{G}^{(g)}(I_{\textrm{ID}_j}, v_{(\textrm{ID}_i,k)})$, where $\textrm{ID}_j \in \mathcal{T}(\textrm{ID}_i),\; k \in \{1,2,\ldots,N\}.$
Thus, each driver identity contributes $8 \times N$ cross-reenactments (eight targets, $N$ driving videos). Rendering each instance with all three generators yields $8 \times N\times 3$ cross-reenactments per driver identity.

The resulting \DatabaseName~provides: \textit{(i)} self-reenactments for genuine verification (the same target and driver), \textit{(ii)} cross-reenactments for impersonation (different target and driver), and \textit{(iii)} generator-aligned triplicates for every target-driver-video tuple. This structure supports a wide range of protocols, including intra-generator and cross-generator testing, cross-dataset generalization, and combined training on multiple generators with evaluation on held-out generators, as we describe next in Sec.~\ref{sec:benchmark}.

\begin{table}
\centering
\renewcommand{\arraystretch}{0.8}
\setlength{\tabcolsep}{4pt}
\begin{tabular}{cccccc}
\textbf{Dataset} & \textbf{Video Split} & \textbf{\#IDs} & \textbf{Self} & \textbf{Cross} & \textbf{Total} \\
\hline

\multirow{3}{*}{\textbf{CREMA-D}} 
& \multicolumn{1}{r}{Development videos per generator} & 61 & 4,392 & 8,280 & 12,672 \\
& \multicolumn{1}{r}{Evaluation videos per generator}  &24 & 1,728 & 3,438 & 5,166 \\
& \multicolumn{1}{r}{Total  videos per generator} &85&  6,120 & 11,718 & 17,838 \\
\hline

\multirow{3}{*}{\textbf{RAVDESS}} 
& \multicolumn{1}{r}{Development videos per generator} & 16 & 960 & 1,905 & 2,865 \\
& \multicolumn{1}{r}{Evaluation videos per generator} & 8 & 480 & 840 & 1,320 \\
& \multicolumn{1}{r}{Total  videos per generator} & 24& 1,440 & 2,745 & 4,185 \\
\hline

\multirow{3}{*}{\textbf{TOTAL}} 
& \multicolumn{1}{r}{Development videos per generator} & 77 & 5,352 & 10,185 & 15,537 \\
& \multicolumn{1}{r}{Evaluation videos per generator} & 32 & 2,208 & 4,278 & 6,486 \\
& \multicolumn{1}{r}{Total  videos per generator} & 109& 7,560 & 14,463 & 22,023 \\
\hline

\multicolumn{3}{c}{\textbf{TOTAL (GAGA+LIVE+HUNY)}} & 22,680 & 43,389 & 66,069 \\
\hline
\end{tabular}
\caption{\textbf{Statistics of \DatabaseName}. ``Self'' corresponds to self-reenactments (genuine avatar) and ``Cross'' corresponds to cross-reenactments (impostor avatar).}
\label{tab:avid_stats}
\end{table}

As a result, the proposed \DatabaseName~comprises 66,069 avatar videos, generated from 109 identities across three avatar generators and two source datasets. The distribution of videos across source datasets and reenactment types is summarized in Table~\ref{tab:avid_stats}.

\subsection{Ethical, FAIR and Data Governance Considerations}\label{ssec:ethics}

The proposed \DatabaseName~is constructed exclusively from publicly available source datasets (CREMA-D~\cite{cao2014crema} and RAVDESS~\cite{livingstone2018ryerson}) distributed under licenses that allow research use, redistribution, and derivative processing (ODbL and CC-BY, respectively). Nevertheless, transforming audiovisual recordings into an avatar fingerprinting benchmark introduces additional ethical considerations due to the use of identifiable human facial motion data.

From a data governance perspective, our design follows transparency, accountability, and data minimization principles informed by the GDPR and EU AI Act~\cite{irigoyen26risks}. We only retain metadata necessary for benchmarking and reproducibility, avoiding unnecessary personal information beyond demographic attributes already available in the original datasets or derived for fairness analysis. The benchmark focuses on behavioral biometric research and security evaluation, rather than identity attribution or surveillance applications.

Regarding participant consent and intended use, although the original datasets were not specifically collected for avatar fingerprinting or impersonation-oriented biometric research, their licensing terms allow research reuse and derivative processing. Consequently, our work emphasizes transparency, provenance preservation, and responsible research use aligned with the spirit of the original data collections. The benchmark is released for academic research, and is intended for scientific evaluation and reproducible research. For this reason, we preserve the provenance of the original datasets and document all generation, preprocessing, and evaluation procedures.

The proposed benchmark is also designed with the FAIR principles (Findable, Accessible, Interoperable, and Reusable) in mind. Findability and accessibility are supported through public hosting with persistent identifiers (\ie, \href{https://github.com/BiometricsAI/AVAPrintDB}{GitHub repository} and \href{https://doi.org/10.5281/zenodo.20548978}{DOI-based archival release in Zenodo}) together with openly available protocols, code, and metadata. Interoperability and reusability are promoted through standardized metadata, versioned releases and documented preprocessing pipelines, yielding an extensible design that accommodates future avatar generators and evaluation scenarios.

\section{Proposed Avatar Fingerprinting Benchmark}\label{sec:benchmark}

Beyond the proposed \DatabaseName~itself, we also define a standardized experimental benchmark, including publicly available avatar fingerprinting systems, identity-disjoint development and evaluation splits, predefined verification trial lists and evaluation procedures. This design ensures that different avatar fingerprinting systems can be compared under identical conditions, facilitating reproducible research and fair benchmarking. This section is organized as follows: the details of the proposed experimental protocol are included in Sec.~\ref{ssec:experimental-protocol} while in Sec.~\ref{ssec:systems} we describe the avatar fingerprinting systems considered in the benchmark, including the exploration for the first time of Foundation Models (DINOv2~\cite{oquab2024dinov} and CLIP~\cite{radford2021learning}).

\subsection{Experimental Protocol}\label{ssec:experimental-protocol}

\subsubsection{Database Splits and Pairs Generation}\label{sssec:genuine_impostor_pairs}

First, in order to evaluate the generalizability of avatar fingerprinting systems to unseen subjects, we partition the 109 identities considered in \DatabaseName~into disjoint development and evaluation splits at the identity level (approximately 70\% / 30\%) to avoid identity leakage. Note that these identity splits do not correspond to those used in NVFAIR or the GAGAvatar benchmark. Instead, they are constructed specifically to ensure demographic balance and fair evaluation. 

\begin{table}
\centering
\renewcommand{\arraystretch}{0.8}
\setlength{\tabcolsep}{4pt}
\begin{tabular}{cccccccc}
 & \multicolumn{1}{c}{} & \multicolumn{3}{c}{\textbf{CREMA-D}} & \multicolumn{3}{c}{\textbf{RAVDESS}} \\ 
\cmidrule(lr){3-5} \cmidrule(lr){6-8} 
\multicolumn{2}{c}{\textbf{Soft-biometrics}} & self & cross & \multicolumn{1}{c}{total} & self & cross & \multicolumn{1}{c}{total} \\ 
\hline
\multirow{2}{*}{\textbf{Gender}} & \multicolumn{1}{c}{Female} & 864 & 1,710 & \multicolumn{1}{c}{2,574} & 240 & 420 & \multicolumn{1}{c}{660} \\
 & \multicolumn{1}{c}{Male} & 864 & 1,728 & \multicolumn{1}{c}{2,592} & 240 & 420 & \multicolumn{1}{c}{660} \\ 
\hline
\multirow{4}{*}{\textbf{Ethnicity}} & \multicolumn{1}{c}{African American} & 576 & 1,152 & \multicolumn{1}{c}{1,728} & - & - & \multicolumn{1}{c}{-} \\
 & \multicolumn{1}{c}{Asian} & 288 & 558 & \multicolumn{1}{c}{846} & 120 & 210 & \multicolumn{1}{c}{330} \\
 & \multicolumn{1}{c}{Caucasian} & 576 & 1,152 & \multicolumn{1}{c}{1,728} & 360 & 630 & \multicolumn{1}{c}{990} \\
 & \multicolumn{1}{c}{Hispanic} & 288 & 576 & \multicolumn{1}{c}{864} & - & - & \multicolumn{1}{c}{-} \\ 
\hline
\multirow{3}{*}{\textbf{Age}} & \multicolumn{1}{c}{20-30} & 648 & 1,278 & \multicolumn{1}{c}{1,926} & 420 & 735 & \multicolumn{1}{c}{1,155} \\
 & \multicolumn{1}{c}{31-45} & 936 & 1,872 & \multicolumn{1}{c}{2,808} & 60 & 105 & \multicolumn{1}{c}{165} \\
 & \multicolumn{1}{c}{46-60} & 144 & 288 & \multicolumn{1}{c}{432} & - & - & \multicolumn{1}{c}{-} \\ 
\hline
\multicolumn{2}{c}{\textbf{Total videos per generator}} & 1,728 & 3,438 & \multicolumn{1}{c}{5,166} & 480 & 840 & \multicolumn{1}{c}{1,320} \\
\hline
\end{tabular}
\caption{\textbf{Soft-biometrics distribution of the evaluation split of \DatabaseName}. ``Self" corresponds to self-reenactments (genuine avatar), ``cross'' corresponds to cross-reenactments (impostor avatar).}
\label{tab:avid_stats_eval}
\end{table}

All avatar videos (self- and cross-reenactments) derived from a given identity are assigned exclusively to that identity's split, so that systems must generalize to unseen identities rather than memorize identity-specific patterns. When metadata are available, the split is stratified to keep similar distributions of gender, age ranges, and ethnicity across the evaluation partition. Table~\ref{tab:avid_stats_eval} shows the soft-biometrics distribution for the \textit{evaluation} split. Systems are trained exclusively on the development split identities, and evaluated exclusively on the disjoint evaluation identities.

Performance is measured on a predefined set of enrollment–test pairs (verification trials) built from the evaluation split videos, which include all possible self-reenactments and a subset of cross-reenactments obtained by driving each target identity with eight different source identities. Each pair is labeled as either genuine or impostor:

\begin{itemize}
    \item \emph{Genuine pair}: both enrollment and test videos are self-reenactments driven by the same underlying identity (\ie, same target and same driver identity), but correspond to different video instances.
    \item \emph{Impostor pair}: the enrollment video is a self-reenactment of a given identity, while the test video is a cross-reenactment with the same target identity (same visual appearance) but driven by a different identity.
\end{itemize}

Formally, a verification trial is defined as a pair of avatar videos $\tau = (V^{e}, V^{t})$, where $V^{e}$ is the enrollment video and $V^{t}$ is the test video.  
Let $\textrm{ID}_d(V)$ denote the underlying human identity driving the avatar video $V$ and $\textrm{ID}_t(V)$ denote the target identity (appearance). Each trial is then labeled as $\ell(\tau) =1$ if  $\textrm{ID}_d(V^{e}) = \textrm{ID}_d(V^{t})$ and $ \textrm{ID}_t(V^{e}) = \textrm{ID}_t(V^{t})$ (genuine pair); or  $\ell(\tau) =0$ if  $\textrm{ID}_d(V^{e}) \neq \textrm{ID}_d(V^{t})$ and $ \textrm{ID}_t(V^{e}) = \textrm{ID}_t(V^{t})$ (impostor pair).

In addition, for genuine pairs, the enrollment and test videos correspond to different recordings of the same identity, ensuring that verification is performed across distinct video instances rather than identical clips.
All possible genuine and impostor pairs are generated exhaustively from the avatar videos in the evaluation split under the above constraints. The resulting list of verification trials and labels is released as part of the benchmark in CSV format, ensuring that all methods are evaluated under identical and reproducible conditions. The summary of the genuine/impostor pairs for the final evaluation split is presented in Table~\ref{tab:eval-pairs}.

\subsubsection{Experimental Scenarios}\label{ssec:experimental_scenarios}

Using the above protocol, we define three main experimental scenarios:
\begin{itemize}
    \item \emph{Intra-dataset, intra-generator}: training and evaluation are performed using the same source dataset and avatar generator. This setting measures the upper-bound performance achievable under matched conditions.
    \item \emph{Intra-dataset, cross-generator}: training and evaluation use the same source dataset but different avatar generators. This scenario evaluates robustness to synthesis-pipeline shift.
    \item \emph{Cross-dataset, intra-generator}: training and evaluation use different source datasets while maintaining the same avatar generator. This setting evaluates robustness to source-domain variation (\eg acquisition conditions, demographic biases, \etc).
\end{itemize}

\begin{table}[t]
\centering
\renewcommand{\arraystretch}{0.8}
\setlength{\tabcolsep}{2pt}
\begin{tabular}{lcccc}
 & \textbf{CREMA-D} & \textbf{RAVDESS} & \textbf{Per generator} &
\begin{tabular}[c]{@{}c@{}}
\textbf{Total} \\
\textbf{(GAGA+LIVE+HUNY)}
\end{tabular} \\
\midrule
\textbf{Genuine pairs} & 124,416 & 28,800 & 153,216 & 459,648 \\
\textbf{Impostor pairs} & 247,536 & 50,400 & 297,936 & 893,808 \\
\hline
\end{tabular}
\caption{\textbf{Summary of genuine and impostor pairs provided in our \DatabaseName~benchmark.}}
\label{tab:eval-pairs}
\end{table}

All these experimental scenarios, summarized in Table~\ref{tab:scenario_summary}, allow us to disentangle three important axes of variation in avatar fingerprinting: \textit{identity generalization} (unseen subjects), \textit{generator generalization} (unseen avatar synthesis pipelines), and \textit{dataset generalization} (unseen source domains). This benchmark provides a controlled framework for understanding whether avatar fingerprinting systems learn robust identity-related motion cues or overfit to dataset- or generator-specific artifacts.

\subsubsection{Performance Metrics}\label{ssec:performance_metrics}

Following standard practice in the literature~\cite{prashnani2024avatar,alsherfawi2025decoding,pedrouzo2025really}, we assess avatar fingerprinting performance using the Area Under the Receiver Operating Characteristic Curve (AUC-ROC). 
This metric measures the ability of the system to separate target from non-target trials across decision thresholds, making it suitable for threshold-independent evaluation under moderate class imbalance.

\begin{table}[t]
\centering
\renewcommand{\arraystretch}{0.8}
\setlength{\tabcolsep}{3pt}
\begin{tabular}{lccl}
\multicolumn{1}{c}{\multirow{2}{*}{\textbf{Scenario}}} & \multicolumn{2}{c}{\textbf{Train and Test}} & \multicolumn{1}{c}{\multirow{2}{*}{\textbf{Goal}}} \\
\cline{2-3}
\multicolumn{1}{c}{}                                   & \textbf{Datasets}    & \textbf{Generators}    & \multicolumn{1}{c}{}                               \\ 
\midrule
Intra-Dataset, Intra-Generator                          & Same                & Same                  & Matched setting                \\
Intra-Dataset, Cross-Generator                          & Same                & Different             & Generator robustness           \\
Cross-Dataset, Intra-Generator                          & Different           & Same                  & Dataset robustness             \\ 
\bottomrule
\end{tabular}
\caption{\textbf{Summary of the experimental scenarios considered in the proposed benchmark.}}
\label{tab:scenario_summary}
\end{table}

To quantify statistical uncertainty, confidence intervals are estimated by bootstrap resampling of evaluation trials. Unless otherwise stated, the metrics reported correspond to mean estimates in the evaluation protocol. Due to space constraints, complete confidence interval tables for all experimental settings are provided in our \href{https://github.com/BiometricsAI/AVAPrintDB}{GitHub repository} together with scripts used for statistical estimation and result reproduction. Bootstrapping was used to estimate the 95\% confidence intervals for the reported AUC and $\Delta$AUC values. This non-parametric procedure quantifies uncertainty through 1,000 resampling iterations with replacement over the evaluation trials.

\subsection{Avatar Fingerprinting Systems}\label{ssec:systems}

\subsubsection{Deep Learning Methods}\label{ssec:dl_methods}

To advance in the topic of avatar fingerprinting under realistic synthesis conditions, we require systems that are reproducible, fully specified, and publicly accessible. Although recent studies have been presented in the literature~\cite{prashnani2024avatar,alsherfawi2025decoding}, none of them have released their implementations, training pipelines or pretrained weights, making it impossible to reproduce their results or conduct fair comparative evaluations.

\begin{figure}[t]
    \centering
    \includegraphics[width=0.85\linewidth]{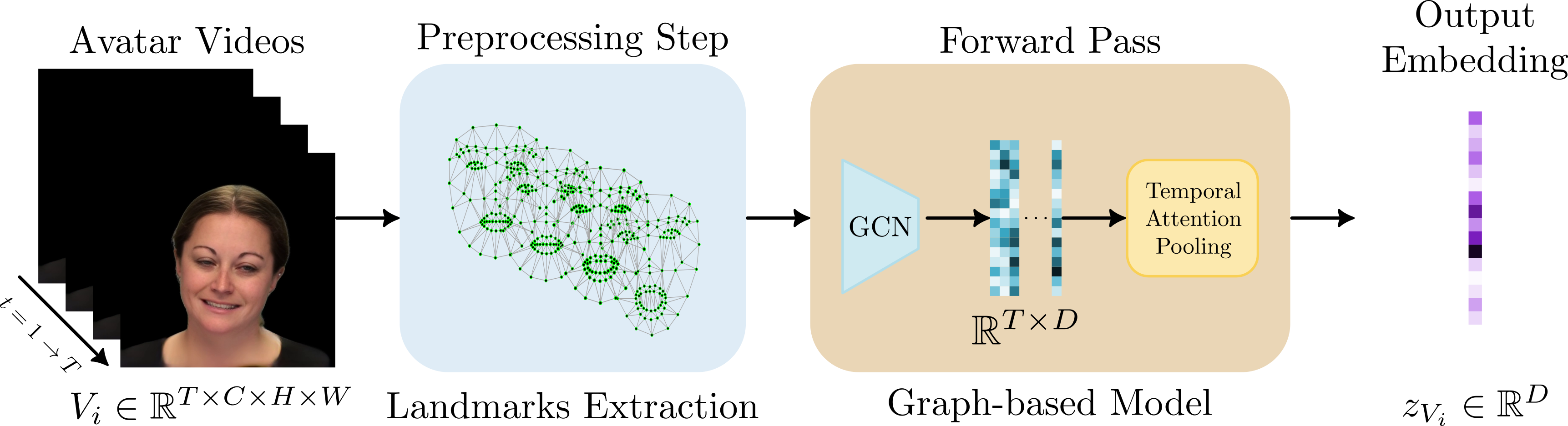}
    \caption{\textbf{Avatar fingerprinting system based on a Graph Convolutional Network (GCN) model}~\cite{pedrouzo2025really}. From an avatar video, landmarks are extracted per frame and used as input for the GCN to obtain a graph embedding per frame. A pooling block generates a final embedding from the graph embeddings.}
    \label{fig:graph-arch}
\end{figure}

For this reason, in the proposed benchmark we evaluate our publicly available graph-based avatar fingerprinting system\footnote{\url{https://github.com/BiDAlab/GAGAvatar-Benchmark}} presented in~\cite{pedrouzo2025really}, which, to the best of our knowledge, remains the only fully reproducible implementation. This system, as depicted in Fig.~\ref{fig:graph-arch}, takes as input 109 facial landmarks per frame in an avatar video, structured as graphs. A Graph Convolutional Network (GCN) generates a graph embedding for each graph (frame) and the aggregation mechanism (temporal attention pooling) yields a single embedding for the video. By evaluating the only publicly available avatar fingerprinting system on a more diverse dataset, our aim is to provide insights into the robustness and limitations of current technology and to establish a foundation for future comparative studies once additional methods become publicly accessible.

\begin{figure}[t]
    \centering
    \includegraphics[width=0.85\linewidth]{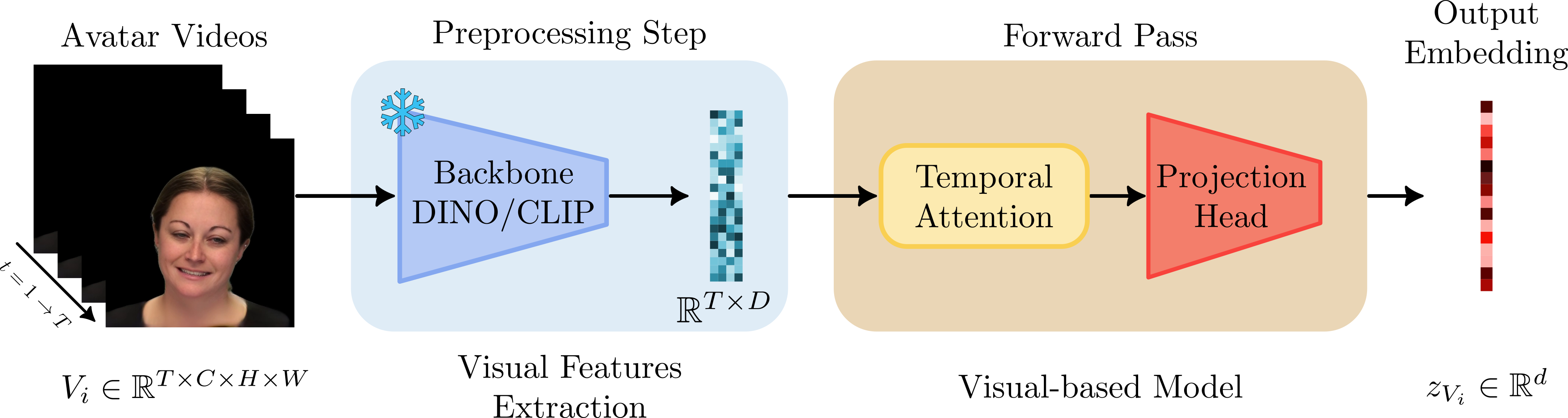}
    \caption{\textbf{Avatar fingerprinting system based on Foundation Models.} The pretrained models (DINOv2~\cite{oquab2024dinov} and CLIP~\cite{radford2021learning}) are frozen and only used to extract visual features for each of the frames. The embeddings obtained are then aggregated via a temporal attention block and a projection head.}
    \label{fig:foundational-arch}
\end{figure}

In addition, we explore for the first time the use of public Foundational Models and attention-based neural networks for avatar fingerprinting. Foundation Models provide a complementary perspective, analyzing whether general-purpose visual representations can capture behavioral biometric cues in avatar videos. Concretely, we use DINOv2~\cite{oquab2024dinov} and CLIP~\cite{radford2021learning} as frame-level feature extractors. Fig.~\ref{fig:foundational-arch} provides a graphical representation of the proposed avatar fingerprinting system. With an input video $V_i \in \mathbb{R}^{T\times C\times H\times W}$ ---where $T$ is the number of frames, $C$ is the number of channels for each frame (RGB), $H$ is the height of the frame, and $W$ is the width of the frame--- the backbone generates a sequence of descriptors $X_i \in \mathbb{R}^{T\times D}$, where $D$ is the embedding dimension specified by DINOv2 or CLIP. In addition to the features, we also propose a multi-head temporal attention module. This module can assign importance weights over time and compress the sequence into a single global representation $\hat{Z_i} \in \mathbb{R}^{D}$. The resulting descriptor is subsequently projected into a compact latent space, $z_{Vi} \in \mathbb{R}^{d}$, where $d < D$. Finally, the embedding is optimized using a triplet loss, following an optimization procedure similar to that proposed in~\cite{prashnani2024avatar, pedrouzo2025really}.

Finally, for completeness, we explore the integration of the GCN-based model with the two foundation models through mean score fusion. This strategy is intended to explore the complementary cues extracted from the different domains and their joint contribution to the final decision.
All avatar fingerprinting methods presented in this study are publicly available in our GitHub\footnotemark[\getrefnumber{fn:github}] repository for reproducibility.

\subsubsection{Implementation and Reproducibility Details}\label{sssec:implementation}

To facilitate reproducibility of the evaluated systems, we release on \href{https://github.com/BiometricsAI/AVAPrintDB}{GitHub} the implementation details together with the benchmark, including preprocessing procedures, configuration files, software dependencies, hardware setup, and pretrained checkpoints.

Regarding preprocessing, the landmark-based graph model follows the original implementation~\cite{pedrouzo2025really}, including face detection, landmark extraction, normalization, and temporal window generation prior to graph construction. For the foundation-model approaches (DINOv2 and CLIP), video frames are resized and normalized according to the corresponding pretrained model specifications before temporal aggregation.

\subsubsection{Video-to-Video Scoring}\label{sssec:video_video}

\begin{figure}[t]
    \centering
    \includegraphics[width=\linewidth]{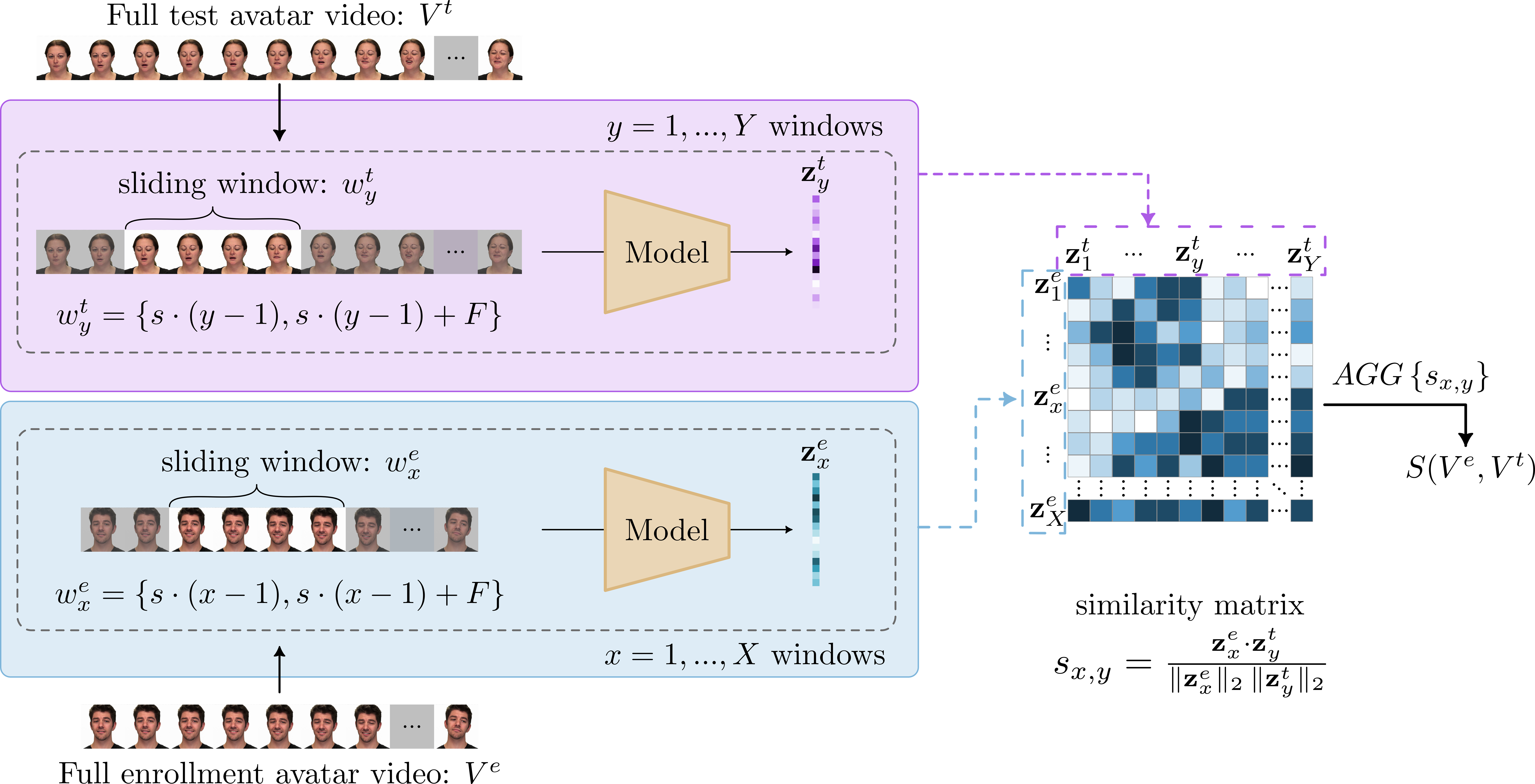}
    \caption{\textbf{Video-to-video scoring for an enrollment-test pair of avatar videos}. Each video is divided into fixed-length windows, with potentially different numbers of windows for $X$ and $Y$. The avatar fingerprinting system generates one embedding $z$ per window, computes the similarity matrix $s_{x,y}$ across all embeddings and averages its values to obtain the final score $S(V^{e},V^{t})$.}
    \label{fig:exp-protocol}
\end{figure}

Since enrollment and test videos may contain different numbers of frames, the proposed avatar fingerprinting systems compute the final verification score at the video-pair level using a window-based embedding and score aggregation procedure (Fig.~\ref{fig:exp-protocol}).

Let the enrollment video be $V^{e}$ and the test video be $V^{t}$. We fix a temporal window length of $F$ frames and a stride of $s=F/2$ frames. Each video is partitioned into all possible windows of length $F$ with stride $s$. 
The choice of window length $F$ determines the temporal context available to capture behavioral patterns such as articulation dynamics, blinking frequency, or characteristic head movements. Trailing frames that do not fill a complete window are discarded. Formally, this yields $\mathcal{W}(V^{e}) = \{w^{e}_1, \dots, w^{e}_X\}$ and $\mathcal{W}(V^{t}) = \{w^{t}_1, \dots, w^{t}_Y\}$
where $w^{z}_i$ are the input data in a time window, $X$ and $Y$ are the number of valid windows in the enrollment $e$ and test $t$ videos, respectively.
Each window is independently processed by the avatar fingerprinting system to obtain a fixed-dimensional embedding $\mathbf{z}^{e}_x = f(w^{e}_x)$ for $x=1,\dots,X$ and $\mathbf{z}^{t}_y = f(w^{t}_y)$ for $y=1,\dots,Y$, where $f(\cdot)$ denotes the learned embedding function.

We then compute the similarity between all enrollments $\mathbf{z}^{e}_x$ and test embeddings $\mathbf{z}^{t}_y$. In our avatar fingerprinting implementation, this is done using cosine similarity.
This produces a $X \times Y$ similarity matrix. The final verification score $S(V^{e},V^{t})$ for the video pair is obtained by averaging all pairwise similarities.
Thus, each enrollment-test pair is assigned a single scalar score that summarizes the average embedding discrepancy across all temporal segments from both videos. 

\section{Experimental Results}\label{sec:experiments}

\subsection{Intra-Dataset and Intra-Generator Scenario}\label{ssec:intra_intra}

We first evaluate avatar fingerprinting systems in the most controlled setting, where development and evaluation use avatars generated from the same dataset and avatar generator. This allows assessing the ability of the systems to capture identity-related motion dynamics from synthetic avatars. Table~\ref{tab:intra} reports AUC (\%) for CREMA-D and RAVDESS using GAGAvatar (GAGA), LivePortrait (LIVE), and HunyuanPortrait (HUNY), together with the ``Fusion'' approach based on aggregation of mean scores.

Across all models and generators, performance is consistently higher on CREMA-D than on RAVDESS (\eg, Fusion achieves 93.8\% vs 83.0\% on GAGA). This is likely due to the larger number of identities and videos in CREMA-D, which provides more variability for learning identity-discriminative motion patterns.

The choice of avatar generator clearly affects performance. LIVE generally produces the highest AUC values, followed by GAGA and HUNY (\eg, for the CREMA-D scenario, the Fusion approach achieves 95.2\% AUC for LIVE versus 93.8\% and 87.6\% for GAGA and HUNY, respectively). This may reflect that differences in motion transfer and synthesis artifacts influence how identity-related motion cues are preserved. 

\begin{table}[t]
\centering
\renewcommand{\arraystretch}{0.8}
\setlength{\tabcolsep}{3pt}
\begin{tabular}{cccccc}
\multirow[c]{2}{*}{} & \multirow[c]{2}{*}{\rotatebox[origin=c]{90}{\textbf{Gen.}}} & \multicolumn{4}{c}{\textbf{Models}} \\
\cline{3-6}
 & & \textbf{Graph} & \textbf{DINOv2} & \textbf{CLIP} & \textbf{Fusion} \\
\hline
\multirow[c]{3}{*}{\rotatebox[origin=c]{90}{C}}
& G 
& 87.8 [87.7, 88.0] & 87.0 [86.9, 87.2] & 86.4 [86.2, 86.5] & \textbf{93.8 [93.7, 93.9]} \\

& L 
& 90.8 [90.7, 91.0] & 88.8 [88.7, 89.0] & 88.6 [88.5, 88.8] & \textbf{95.2 [95.1, 95.3]} \\

& H
& 82.6 [82.5, 82.8] & 79.8 [79.6, 80.0] & 81.0 [80.8, 81.2] & \textbf{87.6 [87.4, 87.7]} \\
\cline{2-6}

\multirow[c]{3}{*}{\rotatebox[origin=c]{90}{R}}
& G
& 75.7 [75.3, 76.0] & 75.9 [75.6, 76.2] & 76.0 [75.7, 76.3] & \textbf{83.0 [82.7, 83.3]}\\

& L
& 76.5 [76.1, 76.8] & 67.2 [66.8, 67.6] & 70.2 [69.8, 70.6] & \textbf{79.4 [79.1, 79.8]} \\

& H
& 72.3 [71.9, 72.6] & 74.0 [73.7, 74.4] & 77.0 [76.7, 77.4] & \textbf{78.8 [78.5, 79.2]} \\

\hline
\end{tabular}%
\caption{\textbf{Intra-dataset and intra-generator scenario}. Reported values correspond to AUC (\%) with uncertainty quantified through bootstrap resampling of evaluation trials. For each generator, we show the AUC, and the corresponding 95\% confidence interval in brackets. Best result is highlighted in \textbf{bold}. G = GAGAvatar, L = LivePortrait, H = HunyuanPortrait, C=CREMA-D, and R=RAVDESS.}
\label{tab:intra}
\end{table}

Among individual systems, the Graph system consistently achieves the best results. In CREMA-D, it reaches 87.8\%, 90.8\%, and 82.6\% AUC for GAGA, LIVE, and HUNY, respectively. However, avatar fingerprinting systems based on the use of Foundation Models (DINOv2 and CLIP) also achieve competitive performance (\eg, GAGA = 87.0\%, LIVE = 88.8\% and HUNY = 79.8\% for the DINOv2 approach), indicating that large-scale pretrained visual representations can also capture identity-discriminative motion information. Similar trends are observed in RAVDESS. Despite relying on score averaging, Fusion consistently achieves the best results, showing that Foundation Models and the Graph system provide complementary information.

\begin{figure}[t]
    \centering
    \includegraphics[width=\linewidth]{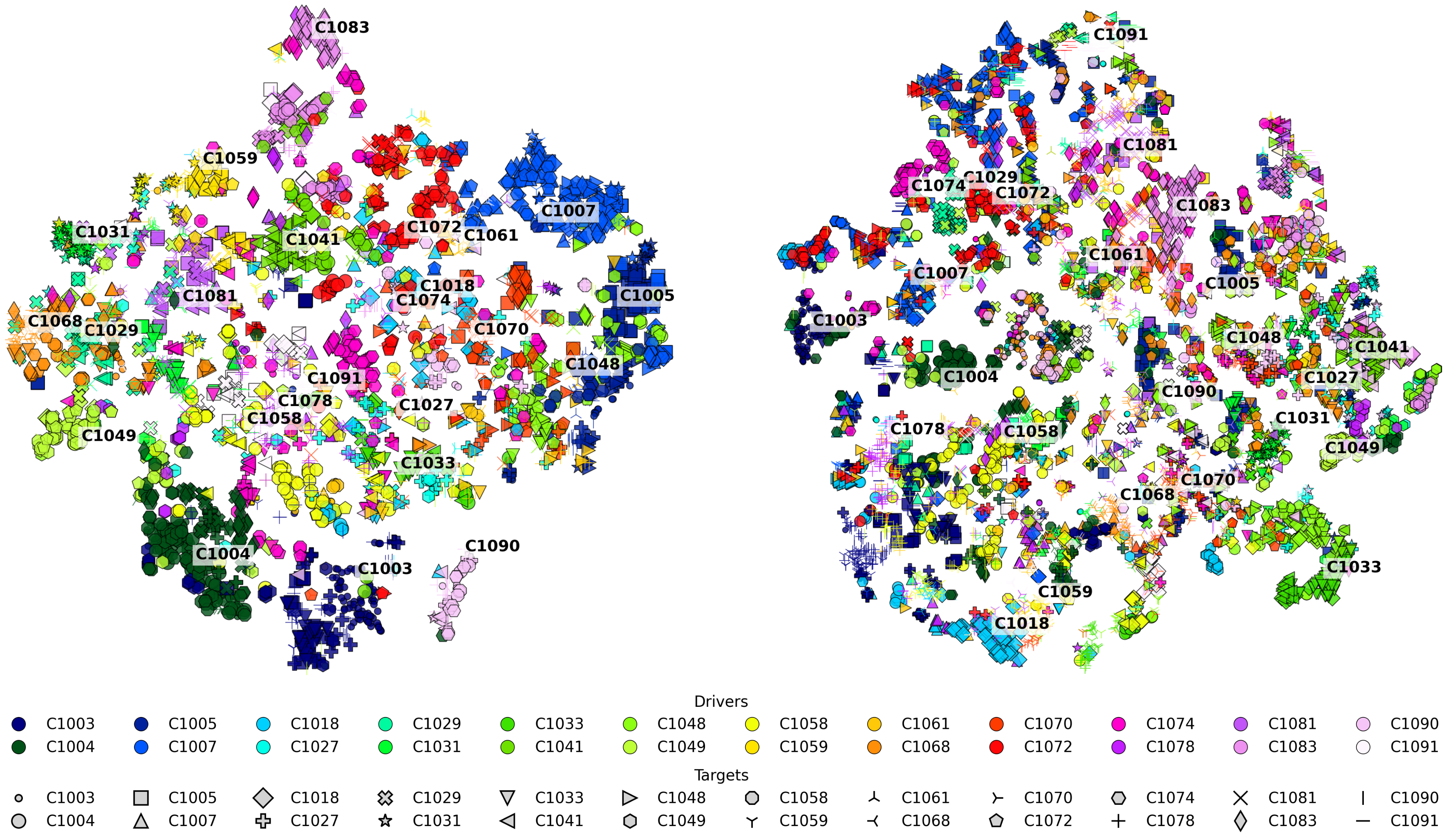}
    \caption{\textbf{Intra-dataset intra-generator t-SNE embedding visualizations}. All obtained from CREMA-D test avatar videos generated with LivePortrait~\cite{guo2024liveportrait} generator. Colors indicate the driving identity and marker types indicate the target identity. \textit{Left}: embeddings obtained with Graph-based system. 
    \textit{Right}: embeddings obtained with DINOv2 system. 
    }
    \label{fig:tsne}
\end{figure}

To further analyze the learned representations, we compute the silhouette score to assess clustering trends according to the driving and target identities. For CREMA-D avatar videos generated with LIVE, the Graph-based model achieves scores of 0.0463 and -0.0759 when clustering by driving and target identity, respectively, while DINOv2 obtains -0.0863 and 0.0702. 
Fig.~\ref{fig:tsne} shows the corresponding visualizations of t-SNE. The embeddings reveal some clustering trends according to the driving identity, although partial overlap remains between some clusters. In addition, the positive DINOv2 score for target identity suggests that target appearance remains partially encoded in foundation-model representations.

\subsection{Intra-Dataset and Cross-Generator Scenario}\label{ssec:intra_cross}

Next, we evaluate the robustness of avatar fingerprinting systems against avatar generators. Specifically, systems are trained on avatars synthesized by one generator and evaluated on avatars synthesized by another, while keeping the source dataset fixed. This setting is particularly relevant in practice, as deployed systems should be robust against future avatar generators. To isolate this effect, we performed intra-dataset, cross-generator experiments. The results are reported in Table~\ref{tab:intergen} as absolute changes in AUC ($\Delta$AUC, expressed as percentage points, not as relative percentages) with respect to the corresponding reference intra-generator.

\begin{table}[t]
\centering
\renewcommand{\arraystretch}{0.8}
\setlength{\tabcolsep}{2pt}
\begin{tabular}{cccccccccc}
\multicolumn{2}{c}{\textbf{Generator}} &
\multicolumn{4}{c}{\textbf{CREMA-D}} &
\multicolumn{4}{c}{\textbf{RAVDESS}} \\
\cmidrule(lr){1-2} \cmidrule(lr){3-6} \cmidrule(lr){7-10}

\textbf{Dev.} & \textbf{Eval.} &
\textbf{Graph} & \textbf{DINOv2} & \textbf{CLIP} & \textbf{Fusion} & 
\textbf{Graph} & \textbf{DINOv2} & \textbf{CLIP} & \textbf{Fusion} \\ 
\hline

GAGA & GAGA & 
\textbf{87.8} & \textbf{87.0} & \textbf{86.4} & \textbf{93.8} & 
\textbf{75.7} & \textbf{75.9} & \textbf{76.0} & \textbf{83.0} \\
GAGA & LIVE & 
-3.2 & -18.2 & -10.6 & -7.8 & 
-8.9 & -11.2 & -10.5 & -12.6 \\
GAGA & HUNY & 
-3.4 & -20.4 & -14.6 & -8.4 & 
-2.6 & -13.7 & -10.8 & -11.5 \\ 
\hline

LIVE & LIVE & 
\textbf{90.8} & \textbf{88.8} & \textbf{88.6} & \textbf{95.2} &
\textbf{76.5} & \textbf{67.2} & \textbf{70.2} & \textbf{79.4} \\
LIVE & GAGA & 
-14.3 & -18.9 & -17.3 & -16.8 &
-7.4 & -1.8 & -7.3 & -10.1 \\
LIVE & HUNY &
-16.9 & -14.6 & -15.0 & -15.9 &
-7.4 & -4.0 & -4.3 & -11.9 \\ 
\hline

HUNY & HUNY &
\textbf{82.6} & \textbf{79.8} & \textbf{81.0} & \textbf{87.6} &
\textbf{72.3} & \textbf{74.0} & \textbf{77.0} & \textbf{78.8} \\ 
HUNY & GAGA &
+2.2 & -16.3 & -8.9 & -8.5 &
-1.9 & -10.2 & -9.9 & -5.8 \\
HUNY & LIVE &
+1.9 & 0.0 & -5.8 & -0.9 &
-4.1 & -6.5 & -7.5 & -4.3 \\
\hline
\end{tabular}%
\caption{\textbf{Intra-dataset, cross-generator scenario}. Reference experiments (intra-generator) are highlighted in \textbf{bold}. Results are shown in AUC (\%) for the reference experiments, and in $\Delta$AUC (percentage points) relative to the corresponding reference experiment for each cross-generator case. A positive deviation means improvement over the reference experiment. Abbreviations: Dev. = Development, and Eval. = Evaluation.}
\label{tab:intergen}
\end{table}

In general, when the avatar generator used for evaluation differs from that used for development, performance drops, indicating limited generator invariance. 
As in Sec.~\ref{ssec:intra_intra}, the Fusion approach consistently achieves the best performance across most configurations, although all systems are affected by generator shifts.

When training in GAGA, the Graph system shows the strongest cross-generator robustness, with relatively small drops in CREMA-D ($-3.2$ and $-3.4$ AUC points when evaluated in LIVE and HUNY, respectively). In contrast, the DINOv2 and CLIP systems are more affected, reaching up to $-20.4$ and $-14.6$, respectively. A similar trend is observed on RAVDESS, although the degradation is generally smaller.

Training in LIVE leads to the poorest transferability. In CREMA-D, all systems experience substantial degradation, with drops reaching $-18.9$ for DINOv2 and $-16.8$ for Fusion. Similar behavior appears on RAVDESS, indicating a stronger generator dependency, possibly from overfitting.

When training on HUNY, the results are more mixed. In some cases, the Graph system even improves when evaluated on other generators (\eg, $+2.2$ on CREMA-D with GAGA, going from 82.6\% AUC to 84.8\% AUC). However, this effect is not consistent between models, as DINOv2 and CLIP show notable degradations.

In general, generalization across avatar generators remains a major challenge. The magnitude and asymmetry of the drops show that current systems are strongly influenced by generator-specific properties.

\subsection{Cross-Dataset and Intra-Generator Scenario}\label{ssec:cross_intra}

We now evaluate the robustness to dataset-induced domain shifts while keeping the avatar generator fixed. In this setting, systems are trained on one dataset (CREMA-D or RAVDESS) and evaluated on the other, allowing us to isolate the impact of variations in data distribution, recording conditions, and subject diversity. The results are reported in Table~\ref{tab:intragen} as AUC (\%) for the reference within the database, and  $\Delta$AUC in percentage points (with respect to this reference) for the cross-dataset experiments.

Unlike the cross-generator scenario (Sec.~\ref{ssec:intra_cross}), cross-dataset transfer shows a strong asymmetry. Training in CREMA-D and testing in RAVDESS consistently degrades performance, whereas the reverse direction  often maintains or even improves over the intra-dataset baseline, indicating sensitivity to dataset-specific characteristics.

\begin{table}[t]
\centering
\renewcommand{\arraystretch}{0.7}
\setlength{\tabcolsep}{3pt}
\setlength{\arrayrulewidth}{0.15pt}

\newcommand{\dshift}[2]{%
\rotatebox[origin=c]{90}{%
\parbox{1.6cm}{\centering
\textbf{#1}$\rightarrow$\textbf{#2}
}
}%
}
\newcommand{\aucshift}[5]{%
\begin{tabular}{@{}c@{}}
\textbf{#1}$\rightarrow$#2 (#3)
\end{tabular}%
}
\begin{tabular}{cccccc}
\rotatebox[origin=c]{90}{\textbf{D$\rightarrow$E}} & \rotatebox[origin=c]{90}{\textbf{Gen.}} & \textbf{Graph} & \textbf{DINOv2} & \textbf{CLIP} & \textbf{Fusion} \\
\hline

\multirow[c]{2}{*}{\dshift{C}{R}}
& G
& \begin{tabular}{@{}c@{}}\aucshift{87.8}{77.3}{-10.5}{-10.9}{-10.2}\end{tabular}
& \begin{tabular}{@{}c@{}}\aucshift{87.0}{79.5}{-7.6}{-8.0}{-7.2}\end{tabular}
& \begin{tabular}{@{}c@{}}\aucshift{86.4}{77.3}{-9.1}{-9.4}{-8.7}\end{tabular}
& \begin{tabular}{@{}c@{}}\aucshift{93.8}{84.2}{-9.6}{-9.9}{-9.4}\end{tabular} \\

& L
& \begin{tabular}{@{}c@{}}\aucshift{90.8}{69.2}{-21.6}{-22.0}{-21.2}\end{tabular}
& \begin{tabular}{@{}c@{}}\aucshift{88.8}{66.9}{-21.9}{-22.3}{-21.5}\end{tabular}
& \begin{tabular}{@{}c@{}}\aucshift{88.6}{59.7}{-28.9}{-29.2}{-28.4}\end{tabular}
& \begin{tabular}{@{}c@{}}\aucshift{95.2}{67.5}{-27.7}{-28.1}{-27.3}\end{tabular} \\

& H
& \begin{tabular}{@{}c@{}}\aucshift{82.6}{74.4}{-8.2}{-8.6}{-7.9}\end{tabular}
& \begin{tabular}{@{}c@{}}\aucshift{79.8}{72.6}{-7.2}{-7.6}{-6.8}\end{tabular}
& \begin{tabular}{@{}c@{}}\aucshift{81.0}{75.6}{-5.4}{-5.8}{-5.0}\end{tabular}
& \begin{tabular}{@{}c@{}}\aucshift{87.6}{77.7}{-9.9}{-10.2}{-9.5}\end{tabular} \\
\hline

\multirow[c]{2}{*}{\dshift{R}{C}}
& G
& \begin{tabular}{@{}c@{}}\aucshift{75.7}{83.5}{+7.9}{+7.6}{+8.3}\end{tabular}
& \begin{tabular}{@{}c@{}}\aucshift{75.9}{80.9}{+5.0}{+4.6}{+5.3}\end{tabular}
& \begin{tabular}{@{}c@{}}\aucshift{76.0}{79.4}{+3.4}{+3.0}{+3.7}\end{tabular}
& \begin{tabular}{@{}c@{}}\aucshift{83.0}{89.3}{+6.3}{+6.1}{+6.6}\end{tabular} \\

& L
& \begin{tabular}{@{}c@{}}\aucshift{76.5}{82.8}{+6.3}{+5.9}{+6.6}\end{tabular}
& \begin{tabular}{@{}c@{}}\aucshift{67.2}{71.5}{+4.3}{+3.9}{+4.8}\end{tabular}
& \begin{tabular}{@{}c@{}}\aucshift{70.2}{76.4}{+6.2}{+5.8}{+6.6}\end{tabular}
& \begin{tabular}{@{}c@{}}\aucshift{79.4}{84.7}{+5.3}{+4.9}{+5.6}\end{tabular} \\

& H
& \begin{tabular}{@{}c@{}}\aucshift{72.3}{77.9}{+5.6}{+5.2}{+6.1}\end{tabular}
& \begin{tabular}{@{}c@{}}\aucshift{74.0}{72.2}{-1.8}{-2.2}{-1.4}\end{tabular}
& \begin{tabular}{@{}c@{}}\aucshift{77.0}{76.6}{-0.4}{-0.9}{-0.1}\end{tabular}
& \begin{tabular}{@{}c@{}}\aucshift{78.8}{83.1}{+4.3}{+3.9}{+4.6}\end{tabular} \\
\hline
\end{tabular}

\caption{\textbf{Cross-dataset, intra-generator scenario}. Each block compares the intra-dataset reference performance against the corresponding cross-dataset evaluation. For each generator and model, we report $\textbf{AUC}_{\mathrm{ref}}\rightarrow \textrm{AUC}_{\mathrm{cross}}$, with the reference AUC (reported in \%) is highlighted in bold, and $\Delta$AUC (reported in percentage points) in parenthesis. D=Development dataset, E=Evaluation dataset, R=RAVDESS, C=CREMA-D, G=GAGAvatar, L=LivePortrait, H=HunyuanPortrait.}
\label{tab:intragen}
\end{table}

For CREMA-D$\rightarrow$RAVDESS (train in CREMA-D, evaluate in RAVDESS), all models show substantial degradation. The Graph system drops by $-10.5$, $-21.6$, and $-8.2$ AUC for GAGA, LIVE, and HUNY, respectively. Larger losses are observed for the DINOv2, CLIP, and Fusion systems in the LIVE setting, where the drops reach $-28.9$ (CLIP) and $-27.7$ (Fusion). These results indicate a limited transferability of the representations learned in CREMA-D. 

In contrast, RAVDESS$\rightarrow$CREMA-D generally improves performance. For example, the Graph system improves by $+7.9$, $+6.3$, and $+5.6$ AUC points for GAGA, LIVE, and HUNY, respectively. Although some configurations show small degradations, the overall trend indicates that the representations learned on RAVDESS are transferred effectively to the more diverse CREMA-D dataset.

The impact of dataset shift also depends on the avatar generator. LIVE consistently exhibits the largest degradation under CREMA-D $\rightarrow$RAVDESS transfer whereas HUNY appears more stable across datasets, and GAGA shows intermediate behavior.

Across systems, Graph is the most robust to dataset shift, while DINOv2 and CLIP are more sensitive, particularly in CREMA-D$\rightarrow$RAVDESS. 
The Fusion approach, although best in the matched setting, remains affected by dataset shift.

Overall, these results show that dataset-induced domain shift is a major challenge for avatar fingerprinting and reinforce the importance of multi-dataset benchmarking.

\subsection{Fairness Analysis}\label{ssec:fairness}

This section provides a preliminary fairness analysis of the selected systems using the soft-biometric annotations available in \DatabaseName~(\ie, gender, ethnicity, and age). 
Results are reported in Tables~\ref{tab:fairness-crema-models} and~\ref{tab:fairness-ravdess-models} (for CREMA-D and RAVDESS, respectively) in terms of AUC (\%) for the intra-dataset and intra-generator scenario.

\begin{table}[t]
\centering
\renewcommand{\arraystretch}{0.8}
\setlength{\tabcolsep}{2.8pt}

\newcommand{\faircell}[3]{#1\,/\,#2~(-#3)}

\begin{tabular}{cccccc}
 & \rotatebox[origin=c]{90}{\textbf{Gen.}}
& \textbf{Graph} & \textbf{DINOv2} & \textbf{CLIP} & \textbf{Fusion} \\
\midrule

\multirow{3}{*}{\rotatebox[origin=c]{90}{\textbf{Gender}}}
& G
& \faircell{88.3}{M}{0.9}
& \faircell{87.0}{M}{3.1}
& \faircell{86.6}{F}{2.2}
& \faircell{94.2}{M}{1.3} \\

& L
& \faircell{91.2}{F}{2.5}
& \faircell{89.0}{F}{9.2}
& \faircell{89.3}{F}{4.9}
& \faircell{95.5}{F}{2.6} \\

& H
& \faircell{83.2}{M}{5.5}
& \faircell{80.2}{F}{8.0}
& \faircell{81.0}{F}{5.1}
& \faircell{87.8}{M}{1.8} \\

\midrule

\multirow{3}{*}{\rotatebox[origin=c]{90}{\textbf{Ethnicity}}}
& G
& \faircell{87.4}{As.}{5.1}
& \faircell{87.1}{A.A.}{5.9}
& \faircell{87.4}{A.A.}{10.4}
& \faircell{93.8}{A.A.}{3.5} \\

& L
& \faircell{90.2}{His.}{4.8}
& \faircell{88.2}{His.}{6.7}
& \faircell{89.2}{A.A.}{2.6}
& \faircell{94.8}{His.}{5.0} \\

& H
& \faircell{81.1}{As.}{9.3}
& \faircell{79.2}{His.}{9.6}
& \faircell{80.9}{As.}{9.8}
& \faircell{86.5}{His.}{6.5} \\

\midrule

\multirow{3}{*}{\rotatebox[origin=c]{90}{\textbf{Age}}}
& G
& \faircell{82.6}{46-60}{19.6}
& \faircell{83.0}{46-60}{16.9}
& \faircell{85.3}{46-60}{5.3}
& \faircell{90.3}{46-60}{13.0} \\

& L
& \faircell{88.8}{46-60}{7.4}
& \faircell{90.8}{20-30}{8.0}
& \faircell{89.6}{20-30}{5.7}
& \faircell{95.5}{31-45}{0.9} \\

& H
& \faircell{81.0}{46-60}{8.3}
& \faircell{80.5}{20-30}{2.2}
& \faircell{77.9}{46-60}{13.8}
& \faircell{85.7}{46-60}{8.4} \\

\bottomrule
\end{tabular}

\caption{\textbf{Fairness summary for CREMA-D avatars}. Each cell reports mean AUC across the subgroups of a given soft-biometric attribute, followed by the worst-performing subgroup and the performance gap between the best and worst subgroup: mean AUC / worst subgroup (gap). Gaps close to zero indicate more stable performance across subgroups. Gen. = avatar generator, G = GAGA, L = LIVE, H = HUNY, F = Female, M = Male, A.A. = African American, As. = Asian, Cau. = Caucasian, and His. = Hispanic.}
\label{tab:fairness-crema-models}
\end{table}

\begin{table}[t]
\centering
\renewcommand{\arraystretch}{0.8}
\setlength{\tabcolsep}{2.8pt}

\newcommand{\faircell}[3]{#1\,/\,#2~(-#3)}

\begin{tabular}{cccccc}
 & \rotatebox[origin=c]{90}{\textbf{Gen.}}
& \textbf{Graph} & \textbf{DINOv2} & \textbf{CLIP} & \textbf{Fusion} \\
\toprule

\multirow{3}{*}{\rotatebox[origin=c]{90}{\textbf{Gender}}}
& G
& \faircell{75.6}{M}{5.3}
& \faircell{76.0}{F}{6.3}
& \faircell{76.7}{M}{3.5}
& \faircell{83.0}{M}{1.2} \\

& L
& \faircell{76.5}{M}{11.3}
& \faircell{67.5}{M}{8.9}
& \faircell{72.4}{M}{8.0}
& \faircell{79.3}{M}{11.3} \\

& H
& \faircell{73.4}{M}{5.4}
& \faircell{74.0}{F}{4.2}
& \faircell{77.1}{M}{8.1}
& \faircell{79.4}{M}{3.2} \\

\midrule

\multirow{3}{*}{\rotatebox[origin=c]{90}{\textbf{Ethnicity}}}
& G
& \faircell{78.4}{Cau.}{9.8}
& \faircell{77.8}{Cau.}{6.5}
& \faircell{79.5}{Cau.}{10.7}
& \faircell{86.7}{Cau.}{11.3} \\

& L
& \faircell{80.7}{Cau.}{15.8}
& \faircell{66.4}{As.}{4.1}
& \faircell{68.3}{As.}{7.4}
& \faircell{82.2}{Cau.}{10.1} \\

& H
& \faircell{74.7}{Cau.}{8.6}
& \faircell{78.6}{Cau.}{12.7}
& \faircell{78.0}{As.}{0.9}
& \faircell{81.0}{Cau.}{7.6} \\

\midrule

\multirow{3}{*}{\rotatebox[origin=c]{90}{\textbf{Age}}}
& G
& \faircell{76.0}{20-30}{0.6}
& \faircell{80.2}{20-30}{9.8}
& \faircell{71.4}{31-45}{12.7}
& \faircell{82.9}{31-45}{0.6} \\

& L
& \faircell{70.0}{31-45}{17.3}
& \faircell{66.0}{31-45}{3.8}
& \faircell{73.7}{20-30}{6.7}
& \faircell{74.5}{31-45}{12.7} \\

& H
& \faircell{71.6}{31-45}{2.7}
& \faircell{77.5}{20-30}{8.2}
& \faircell{81.6}{20-30}{7.8}
& \faircell{81.6}{20-30}{5.3} \\

\bottomrule
\end{tabular}

\caption{\textbf{Fairness summary for RAVDESS avatars}. Each cell reports mean AUC across subgroups of a soft-biometric attribute, the worst subgroup and the best--worst gap: mean AUC / worst subgroup (gap). Lower gaps indicate more stable subgroups performance. Gen. = avatar generator, G = GAGA, L = LIVE, H = HUNY, F = Female, M = Male, As. = Asian, and Cau. = Caucasian.}
\label{tab:fairness-ravdess-models}
\end{table}

Fairness is evaluated at the verification-trial level. 
To assess demographic effects consistently, we define all subgroup labels (gender, ethnicity, and age) based on the \emph{enrollment identity}, reflecting the intended authentication scenario in which the system verifies whether a query avatar matches a specific user who may be impersonated. 

Gender-related differences are small and inconsistent: some configurations favor male subjects and others favor female subjects, with no systematic bias.

Ethnicity-related variability is more pronounced. On CREMA-D (Table~\ref{tab:fairness-crema-models}) the best subgroup is often Caucasian, while Asian and Hispanic groups tend to score lower (notably for HunyuanPortrait), though this is not uniform across all models, and the African American group is competitive in several cases. RAVDESS shows similar variability, with conclusions limited by its lower diversity. 


Age shows the greatest disparities. 
In CREMA-D, the 46--60 age group is frequently the worst-performing subgroup and yields the largest gaps, while RAVDESS lacks this age range entirely and its two available age ranges vary widely. 

In general, subgroup effects are not systematic but depend on the interaction of the data set, the generator, and the representation, and should be interpreted cautiously given the underrepresentation of certain age and ethnic groups (especially in RAVDESS). Nevertheless, these analyses provide an initial demographic benchmark and motivate future work on more diverse and balanced data.

\subsection{Discussion}\label{ssec:discussion}

The experimental results reveal that avatar fingerprinting is feasible under controlled conditions, but remains highly sensitive to distribution shifts introduced by both avatar generators and source datasets. 

In the intra-dataset, intra-generator setting, all systems achieve strong verification performance, indicating that identity-related motion information is preserved in synthetic avatars. The embedding visualizations further support these findings, showing identity-consistent clustering under matched conditions. Performance, however, varies across generators: LIVE generally yields the strongest matched-condition performance, which means it may produce more stable and temporally consistent motion patterns, which facilitates the extraction of discriminative behavioral features, whereas other generators may introduce smoothing effects or artifacts that partially distort such cues.
At the model level, the Graph system tends to be the strongest standalone approach, indicating that structured motion representations transfer well to avatar verification. Nevertheless, Foundation Models remain competitive, showing that large-scale pretrained visual features also capture identity-discriminative behavioral cues. Their complementarity is further reflected by the consistent gains obtained with Fusion. Performance is also generally higher on CREMA-D than on RAVDESS, suggesting that the composition of the dataset, the diversity of the subject, and the variability of training influence the quality of the behavioral representations learned.

The cross-generator and cross-dataset experiments show that current systems have not yet learned fully invariant behavioral representations. Cross-generator results reveal a strong asymmetry, with some generators (\eg, GAGA) transferring comparatively well and others (particularly LIVE) showing much poorer transferability, indicating partial reliance on generator-specific regularities. The marked asymmetry between CREMA-D$\rightarrow$RAVDESS and RAVDESS$\rightarrow$CREMA-D transfer further suggests that the larger variability of CREMA-D encourages richer but more dataset-dependent representations, while the more controlled structure of RAVDESS encourages stable motion cues that generalize better.

The exploration of Foundation Models highlights their potential for avatar fingerprinting. DINOv2 and CLIP show competitive performance despite not being designed for avatar fingerprinting, while their complementarity with the Graph representation (evidenced by the `Fusion' approach) suggests that combining geometric and pretrained visual features is a promising direction for future behavioral biometrics.

Finally, fairness analysis indicates that demographic effects are not systematic but depend on the interaction between dataset, generator, and representation, reinforcing the importance of balanced benchmarks and standardized evaluation protocols.

\section{Limitations and Conclusion}\label{sec:conclusions}

This article introduces \DatabaseName, a public multi-generator photorealistic talking-head avatar database and benchmark for avatar fingerprinting, comprising 66,069 videos generated across three avatar synthesis paradigms and two audiovisual source datasets. Through evaluation under generator and dataset shift, the benchmark validates avatar fingerprinting as a viable behavioral biometric task while providing a reproducible framework for studying robustness across synthesis pipelines and source domains. Results show that identity-related motion cues persist across synthetic avatars, but current avatar fingerprinting systems are highly sensitive to dataset and generator shifts. This motivates research toward generator- and domain-robust avatar fingerprinting methods.

Despite these findings and the reproducibility advantages of \DatabaseName, some limitations remain. First, the three included generators, though they span distinct paradigms, cannot fully represent the rapidly evolving landscape of avatar synthesis. Extending the benchmark to additional generators is an important future direction. Second, the source datasets (CREMA-D and RAVDESS) are controlled recordings that provide strong control and demographic metadata but do not fully capture real-world conversational variability, unconstrained environments, or deployment conditions. Third, the demographic coverage is imperfect. Certain age and ethnic groups are underrepresented, particularly in RAVDESS, which limits the scope of the analysis.

From a modeling perspective, the benchmark focuses on 1:1 verification rather than identification, impersonator detection, adversarial evaluation, or continuous authentication settings, and the current scoring relies on relatively simple temporal aggregation and score fusion, which may not optimally exploit discriminative temporal segments or long-range behavioral dependencies. More advanced temporal modeling, representation learning, and domain-robust optimization strategies may be required to improve robustness under generator and dataset shift.

To summarize, \DatabaseName~provides a reproducible and extensible benchmark for studying avatar fingerprints under diverse data synthesis conditions. We hope it facilitates the development of more robust and practically useful avatar fingerprinting systems for authentication in avatar-mediated communication.

\section*{Acknowledgments}

This project has been supported by PowerAI+ (SI4/PJI/2024-00062 Comunidad de Madrid and UAM), Cátedra ENIA UAM-Veridas en IA Responsable (NextGenerationEU PRTR TSI-100927-2023-2), M2RAI (PID2024-160053OB-I00 MICIU/FEDER), and TRUST-ID (PID2025-173396OB-I00 MICIU/AEI and the EU).

\section*{Declaration of AI-assisted technologies in the manuscript preparation process}

During the preparation of this work the author(s) used ChatGPT in order to check grammar and spelling. After using this tool, the author(s) reviewed and edited the content as needed and take(s) full responsibility for the content of the published article.

\bibliographystyle{elsarticle-num-names}
\bibliography{references}


\end{document}